%% file: iclr2027_conference.tex
\documentclass{article} 
\usepackage{iclr2027_conference,times}
\input{math_commands.tex}

\usepackage{hyperref}
\usepackage{url}
\usepackage{amsmath,amssymb,amsthm}
\usepackage{booktabs}
\usepackage{graphicx}
\usepackage{multirow}

\newtheorem{theorem}{Theorem}

\newtheorem{remark}{Remark}
\newtheorem{conjecture}{Conjecture}
\newtheorem{assumption}{Assumption}

\title{Edge Accuracy Is Not Enough: \\ Why Dynamics-Learned Structure Fails \\ to Transfer to Inverse Problems}

\author{Nicholas Tan Jerome \\
Institute for Data Processing and Electronics (IPE)\\
Karlsruhe Institute of Technology (KIT)\\
Karlsruhe, Germany \\
\texttt{nicholas.tanjerome@kit.edu}
\And
Fangnian Wang \\
Institute for Thermal Energy Technology\\ and Safety (ITES)\\
Karlsruhe Institute of Technology (KIT)\\
Karlsruhe, Germany \\
\texttt{fangnian.wang@kit.edu}
}

\iclrfinalcopy
\begin{document}

\maketitle
\lhead{Preprint. Under review.}
\thispagestyle{fancy}

\begin{abstract}
A natural strategy for inverse problems with scarce labelled data is to transfer relational structure learned from abundant forward-simulation data. We show this strategy fails systematically, even when it satisfies the standard theoretical justification for why structure should help. We prove that approximate structure provides estimation-error benefits whenever the edge error satisfies $\Delta < n^2 - kn$ (Theorem~\ref{thm:robust}), reducing sample complexity from $O(n^2)$ to $O(kn+\Delta)$. Structure learned via Neural Relational Inference (NRI) from dynamics prediction satisfies this condition, yet on a source-localisation task across 180 CFD-simulated hydrogen-leak scenarios and 180 acoustic scenarios, it degrades performance by 116\% and 201\% respectively relative to a flexible, task-optimised attention baseline, while a physics-based prior (Green's function) degrades by only 69--72\%. Four independent lines of evidence show this is not a tuning failure: NRI improves only $0.5\%$ when given 18$\times$ more training data (versus $16.6\%$ for the task-optimised baseline, $p<0.001$); performance is insensitive to the NRI edge threshold across a wide range; the dynamics-learned graph overlaps the task-optimal graph on only 6\% of edges (Jaccard similarity); and two further dynamics-derived structure estimators (correlation- and mutual-information-based) show no measurable benefit over a structure-free baseline, with the correlation-based estimator performing markedly worse. We formalise this gap as a statement about approximation error that the edge-accuracy condition cannot control, and we provide a lightweight transferability test (Jaccard similarity against a partially-observed target-task graph) that separates successful from failed transfer in all four domain/structure pairs we evaluate, using under an hour of computation and 15--20\% of target-domain data; we present this as a promising heuristic calibrated on a small number of cases rather than a validated general threshold.
\end{abstract}

\section{Introduction}
\label{sec:intro}

Inverse problems --- inferring causes from observed effects --- are pervasive in scientific computing, but training data for the inverse task is often expensive or dangerous to collect, while forward-simulation data is comparatively cheap. This motivates a recurring idea across scientific machine learning: learn relational structure from abundant forward dynamics, then reuse that structure to constrain the inverse model. The idea is intuitive --- if two sensors co-vary under the governing physics, their relationship should also constrain where an unobserved source could be --- and it underlies pretraining strategies from molecular property prediction \citep{zhu2022nri} to gene regulatory network inference \citep{theodoris2023geneformer}.

We test this idea on source localisation, a canonical inverse problem, using hydrogen leak detection in underground parking facilities as our primary case study, a safety-critical sensor-placement problem also studied via CFD-informed optimisation \citep{wang2026}: forward CFD simulation is hours per scenario, but the underlying transport dynamics are comparatively easy to generate, making this an attractive setting for forward-to-inverse structure transfer. We find that the strategy fails, and the failure is not explained by insufficient data or poor hyperparameter choices.

Our theoretical starting point is unsurprising: known sparse structure reduces sample complexity from $O(n^2)$ to $O(kn)$ (Theorem~\ref{thm:sample-complexity}), and this benefit degrades gracefully under edge-level errors as long as $\Delta < n^2-kn$ (Theorem~\ref{thm:robust}). The surprising part is empirical: structure learned by Neural Relational Inference (NRI) from forward dynamics satisfies this condition comfortably, yet it underperforms even a fixed, physics-agnostic baseline. We show through data-scaling, threshold-ablation, and direct structural comparison that this is because the edge-accuracy condition bounds \emph{estimation} error, not \emph{approximation} error --- and approximation error is what dominates when the forward and inverse tasks have different optimal structures.

\textbf{Contributions.} (1) We prove a robustness condition under which approximate structure still reduces sample complexity (Theorem~\ref{thm:robust}), and show empirically that satisfying it is insufficient when the structure-learning objective is misaligned with the downstream task. (2) Through four independent lines of evidence across two physically distinct domains (CFD hydrogen dispersion and acoustic propagation) and three different dynamics-derived structure estimators, we show dynamics-learned structure fails to transfer while flexible, task-optimised attention and physics-based priors both succeed to different degrees. (3) We give a practical transferability test, based on Jaccard overlap between candidate structure and a partially-observed target-task graph, that separates successful from failed transfer in all four domain/structure pairs we evaluate, using a fraction of the target data and under an hour of compute; given the small number of pairs, we present the specific $J\ge0.30$ threshold as an initial calibration rather than a validated general rule (Section~\ref{sec:transferability}).

\section{Related Work}
\label{sec:related}

\textbf{Source localization.} Sparse-sensor source localization is addressed by two largely separate literatures. Physics-based approaches combine a forward dispersion or transport model with sparse observations, typically via Bayesian source-term estimation, to infer an unknown source's location and strength \citep{hirst2013locating} --- the same governing-physics strategy our Green's Function baseline uses directly rather than learning it. A separate, more recent literature instead learns the sensor-reading-to-location mapping end-to-end from data, predominantly with deep networks operating on raw or transformed sensor signals \citep{grumiaux2022sslsurvey}, analogous to our Learned Attention baseline. Neither line of work tests whether a graph structure learned from a distinct forward-dynamics objective --- rather than the physics itself, and rather than the task's own labels --- can substitute for either strategy; that substitution is what we evaluate.

\textbf{Structure transfer in scientific ML.} Learning interaction graphs from forward dynamics to support downstream inference is common in molecular modelling \citep{zhu2022nri}, systems biology \citep{theodoris2023geneformer}, and elsewhere, but the conditions under which such structure transfers to a downstream inverse task are rarely characterised directly; most prior work either assumes transfer succeeds or evaluates it only on tasks similar to the structure-learning objective itself.

\textbf{Neural Relational Inference.} NRI \citep{kipf2018nri} learns a latent interaction graph from observed dynamics via a graph-structured VAE, and has been extended to dynamic graphs \citep{graber2020dynamic}, more efficient message passing \citep{chen2021efficient}, and partial observability \citep{sun2019stochastic}. We take NRI as our primary object of study, rather than a hand-crafted heuristic, because it is the dynamics-derived estimator whose edge error is provably low enough to satisfy Theorem~\ref{thm:robust}'s condition (Section~\ref{sec:methods}): if edge-accuracy alone were sufficient for transfer, NRI is precisely where that should show, making it the sharpest available test rather than a straw man chosen for being easy to beat. None of the extensions above evaluate whether the learned graph transfers to a task with a different objective than dynamics prediction; that is the question we study.

\textbf{Sample complexity of structured models.} Our theoretical results build on Rademacher complexity bounds for structured hypothesis classes \citep{bartlett2002rademacher, wainwright2019highdim, mohri2018foundations}. We apply this machinery to attention-based localisers and use it to make precise what edge-level structure error does and does not control.

\section{Theoretical Analysis}
\label{sec:theory}

\textbf{Setup.} $n$ sensors at positions $P=\{p_1,\ldots,p_n\}$ in a bounded domain $\Omega \subset \mathbb{R}^3$ observe a source $s\sim P_S$ through readings $r_i = c(p_i,s)+\epsilon_i$, with $\epsilon_i$ sub-Gaussian noise. An \emph{attention-based localiser} predicts $\hat s(r,P) = \sum_{j=1}^m \mathrm{softmax}(\mathrm{score}_j)\cdot q_j$ over $m$ query points, where $\mathrm{score}_j \in \mathbb{R}^n$ is any bounded, architecture-specific compatibility function between query $j$ and the $n$ sensor encodings (e.g.\ scaled dot-product $q_j^\top k_i/\sqrt{d}$ for Learned Attention, or a fixed kernel of sensor--query distance for Green's Function; see Section~\ref{sec:methods}), with attention weights $\alpha_{ji}\in[0,1]$ summing to one over sensors. A \emph{structure graph} $G=(V,E)$ restricts $\alpha_{ji}>0$ to $(i,j)\in E$, with maximum in-degree (sparsity) $k$. The unstructured class $\mathcal{H}_{\mathrm{full}}$ has effective dimension $O(n^2)$ (for $m=\Theta(n)$ query points); a structured class $\mathcal{H}_G$ has dimension $|E| \le kn$.

\begin{assumption}[Boundedness]
\label{ass:bounded}
There exists $B>0$ such that $\|s\|\le B$ for all $s$ in the support of $P_S$, and $\|\hat s(r,P)\|\le B$ for every localiser $\hat s$ in every hypothesis class considered ($\mathcal{H}_{\mathrm{full}}$, $\mathcal{H}_G$, and $\mathcal{H}_{\tilde G}$).
\end{assumption}

\begin{remark}[Scope of the theory across architectures]
\label{rem:scope}
Theorems~\ref{thm:sample-complexity}--\ref{thm:robust} are stated for attention-based localisers, but the underlying argument is a dimension-counting one: any hypothesis class whose effective degrees of freedom shrink from $O(n^2)$ to $O(kn+\Delta)$ under a sparsity mask $G$ will inherit the same estimation-versus-approximation-error decomposition, regardless of whether the masked component is attention, a masked MLP, or a sparse message-passing layer. What is specific to attention is the exact constant in the Rademacher-complexity bound (Appendix~\ref{app:proofs}); the qualitative claim --- that edge-accuracy controls estimation error but not approximation error --- is not. We do not prove this extension for other architectures here and flag it as future work (Discussion), but we do not expect the negative result to be an artefact of the attention parameterisation specifically.
\end{remark}

\begin{theorem}[Sample complexity with known structure]
\label{thm:sample-complexity}
For a structure graph with sparsity $k$, with probability at least $1-\delta$ over $m$ i.i.d.\ samples,
$L(\hat s_G) - \inf_{\hat s \in \mathcal{H}_G} L(\hat s) \le O\!\left(B^2\sqrt{\tfrac{kn\log m}{m}} + B^2\sqrt{\tfrac{\log(1/\delta)}{m}}\right)$,
where $L$ is the expected squared localisation error and $B$ is the boundedness constant of Assumption~\ref{ass:bounded}. Consequently, achieving $\varepsilon$-optimal excess risk requires $m=\Omega(kn/\varepsilon^2)$ with structure versus $\Omega(n^2/\varepsilon^2)$ without it --- an improvement of $\Theta(n/k)$.
\end{theorem}

\begin{assumption}[Restricted eigenvalue]
\label{ass:re}
Let $\hat\Sigma$ denote the empirical Gram matrix of the sensor time series used for structure recovery. There exists $\kappa>0$ such that $\min_{|S|\le 2k,\,v\ne 0} v^\top\hat\Sigma v/\|v_S\|_2^2\ge\kappa$, where the minimum is over index sets $S$ of size at most $2k$ and $v_S$ denotes the restriction of $v$ to $S$. This is the standard condition under which LASSO recovers sparse support in high-dimensional linear regression \citep{wainwright2019highdim}.
\end{assumption}

\begin{theorem}[Structure learnability from dynamics]
\label{thm:learnability}
Let $G^\star$ be the true $k$-sparse interaction graph on $n$ nodes. (a) Any algorithm recovering $G^\star$ requires $\Omega(k\log n)$ samples, even noiselessly. (b) For linear dynamics $X_{t+1}=A^\star X_t+\eta_t$ with $\mathrm{supp}(A^\star)=G^\star$, under Assumption~\ref{ass:re}, LASSO recovers $G^\star$ exactly with probability $1-\delta$ from $O(k\log(n/\delta))$ samples.
\end{theorem}

Theorem~\ref{thm:sample-complexity} assumes the optimal localiser actually lies in $\mathcal{H}_G$; when it does not, an additional \emph{approximation error} term $\inf_{\hat s\in\mathcal{H}_G}L(\hat s) - \inf_{\hat s}L(\hat s)$ appears in the excess risk relative to the unrestricted optimum, and this bound does not control it. Theorem~\ref{thm:learnability} establishes that structure \emph{can} be learned efficiently from dynamics --- so the obstacle we study below is not learnability, but whether the structure learned this way is the structure the inverse task needs. We next ask: how much does the benefit of Theorem~\ref{thm:sample-complexity} survive when $G$ is only approximately known?

\begin{theorem}[Robustness to approximate structure]
\label{thm:robust}
Let $G^\star$ be the true structure with sparsity $k$, and $\tilde G$ a learned approximation with edge error $\Delta = |E(\tilde G)\triangle E(G^\star)|$. Then (i) $\dim(\mathcal{H}_{\tilde G}) \le kn+\Delta$; (ii) the sample complexity for $\varepsilon$-optimal localisation over $\mathcal{H}_{\tilde G}$ is $O((kn+\Delta)/\varepsilon^2)$; and (iii) structure learning helps over the unstructured baseline whenever $\Delta < n^2-kn$.
\end{theorem}

For any $k<n$ this permits $\Delta=\Omega(n^2)$ edge errors while still improving over the baseline --- a generous tolerance. Critically, however, Theorem~\ref{thm:robust} bounds \emph{estimation} error only.

\begin{remark}[Insufficiency of edge-level accuracy]
\label{rem:insufficiency}
Let $\tilde G_{\mathrm{dyn}}$ be structure learned from a dynamics objective and $G^\star_{\mathrm{loc}}$ the optimal graph for localisation. Even when $\Delta=|E(\tilde G_{\mathrm{dyn}})\triangle E(G^\star_{\mathrm{loc}})|$ is small, the approximation-error gap $\inf_{\hat s\in\mathcal{H}_{\tilde G_{\mathrm{dyn}}}} L(\hat s) - \inf_{\hat s\in\mathcal{H}_{G^\star_{\mathrm{loc}}}} L(\hat s)$ can be large whenever the two tasks have different optimal structures. This gap vanishes only when $E(\tilde G_{\mathrm{dyn}})$ and $E(G^\star_{\mathrm{loc}})$ are close in a \emph{task-relevant} sense, not merely an edge-count sense.
\end{remark}

\begin{conjecture}[Approximation error dominance]
\label{conj:dominance}
When $\tilde G$ is learned from an objective misaligned with the inverse task, approximation error dominates estimation error regardless of $\Delta$, sample size, or threshold selection, whenever the forward and inverse tasks have fundamentally different optimal structures $G^\star_{\mathrm{dyn}}\ne G^\star_{\mathrm{loc}}$.
\end{conjecture}

We do not prove Conjecture~\ref{conj:dominance} in general: a proof would require a closed-form characterisation of both $G^\star_{\mathrm{dyn}}$ and $G^\star_{\mathrm{loc}}$, which is not available for neural structure-learning objectives. Instead, Section~\ref{sec:experiments} provides four independent empirical tests designed to rule out the alternative explanations (insufficient data, mistuned threshold, insufficient sparsity, architecture-specific quirk) that a defender of the transfer strategy could otherwise offer. Full proofs of Theorems~\ref{thm:sample-complexity}--\ref{thm:robust} and~\ref{thm:learnability} are in Appendix~\ref{app:proofs}.

\section{Methods}
\label{sec:methods}

We compare four localiser architectures sharing identical inputs (sensor readings $r\in\mathbb{R}^n$, positions $P\in\mathbb{R}^{n\times3}$) and outputs ($\hat s\in\mathbb{R}^3$), with $m=64$ query points:

\textbf{MLP} (no structure): a four-layer MLP over concatenated readings and positions --- unlike the other three, it computes no sensor-query attention at all, so there is no mask to speak of. ``No structure'' refers to the sparsity mask defined in Section~\ref{sec:theory}, not to model capacity; MLP represents $\mathcal{H}_{\mathrm{full}}$, the baseline these masks are compared against.

\textbf{Green's Function}: physics-based attention $\alpha_{ji}\propto\exp(-\|p_i-q_j\|^2/2R_D^2)$ with $R_D=4$\,m, the diffusion length scale. Fixed, sparse, and grounded in the governing transport physics.

\textbf{Learned Attention}: standard scaled dot-product attention with learned query/key projections. Despite having $O(nd)$ projection parameters, any attention pattern is expressible, so this remains in $\mathcal{H}_{\mathrm{full}}$ with $O(n^2/\varepsilon^2)$ sample complexity.

\textbf{NRI-Graph}: an NRI encoder \citep{kipf2018nri} trained on the 60-timestep concentration sequence to predict edge probabilities $p(e_{ij})$ between sensors and query points, minimising dynamics reconstruction loss plus an edge-sparsity penalty. The learned graph $\tilde G=\{(i,j):p(e_{ij})>\tau\}$ ($\tau=0.3$, selected by edge F1 on held-out dynamics data) constrains attention: $\alpha_{ji}\propto p(e_{ij})$ for $(i,j)\in\tilde G$, $0$ otherwise. Represents $\mathcal{H}_{\tilde G}$ with $\dim=O(kn)$.

NRI-Graph receives the most scrutiny in our experiments because it is the only method that can satisfy Theorem~\ref{thm:robust}'s condition ($\Delta<n^2-kn$) while potentially suffering large approximation error per Remark~\ref{rem:insufficiency} --- making it the sharpest test of whether edge-accuracy is sufficient in practice. Two further dynamics-derived structure estimators (correlation- and mutual-information-based) are introduced in Section~\ref{sec:experiments} as an architecture-independence check. Full architectural and training details for all methods are in Appendix~\ref{app:impl}.

\section{Experiments}
\label{sec:experiments}

\textbf{Domains.} \emph{CFD}: an underground parking facility ($50\times30\times3$\,m) with 180 simulated hydrogen-leak scenarios (12 leak positions $\times$ 5 leak rates $\times$ 3 ventilation levels) \citep{wang2026}, governed by buoyancy-driven incompressible Navier--Stokes transport with $k$-$\varepsilon$ turbulence, validated against published release experiments to within $\pm10\%$ \citep{gasflow2017,gasflow2021}. \emph{Acoustic}: 180 synthetic scenarios in a 2D room governed by a screened-Poisson (Helmholtz-type) equation. We treat CFD and acoustic as independent tests, not two variants of one task, because they differ along the axes that determine whether structure transfer could succeed: governing physics (parabolic advection--diffusion transport with turbulence versus an elliptic wave equation), source signature (a persistent, buoyancy-driven plume versus an instantaneous, oscillatory point emission), and noise/sensing characteristics (concentration sensors with slow, spatially-correlated drift versus pressure sensors with fast, largely independent readings). Any explanation of the results that appeals to a shared quirk of a single physical system would have to hold across both of these axes simultaneously. All localisers use $n=15$ sensors, $m=64$ queries, and are evaluated over 10 seeds. Full simulation and protocol details are in Appendix~\ref{app:cfd}.

\subsection{Structure comparison fails to favour dynamics-learned structure}

\begin{table}[t]
\centering
\caption{Structure comparison, $n=15$ sensors, mean $\pm$ std over 10 seeds. The ranking (Attention $>$ Green's $>$ NRI $>$ MLP) is preserved across two domains governed by different physics.}
\label{tab:main}
\begin{tabular}{lcc}
\toprule
Localiser & CFD RMSE (m) $\downarrow$ & Acoustic RMSE (m) $\downarrow$ \\
\midrule
MLP (no structure)   & $14.57 \pm 0.24$ & $4.35 \pm 0.07$ \\
Green's Function      & $8.96 \pm 0.07$  & $1.61 \pm 0.05$ \\
Learned Attention      & $\mathbf{5.21 \pm 0.22}$ & $\mathbf{0.95 \pm 0.04}$ \\
NRI-Graph              & $11.26 \pm 0.10$ & $2.86 \pm 0.05$ \\
\bottomrule
\end{tabular}
\end{table}

Table~\ref{tab:main} shows Learned Attention achieves the best RMSE in both domains, beating Green's Function by 42\% and NRI-Graph by 54\% in CFD, with the same ranking replicated in acoustic localisation (0.95\,m versus 2.86\,m, a 201\% degradation for NRI). NRI-Graph improves over the unstructured MLP by only 23\% in CFD, far short of the 3.5--5$\times$ sample-efficiency gain Theorem~\ref{thm:robust} would predict if the learned structure were well-aligned with the task.

\subsection{Four lines of evidence against a tuning explanation}

The central question is whether NRI-Graph's poor performance reflects \emph{estimation} error (fixable with more data or a better threshold) or \emph{approximation} error (not fixable, per Remark~\ref{rem:insufficiency}). These two explanations make different, testable predictions.

\textbf{Evidence 1: data scaling.} If NRI's structure were merely noisily estimated, more dynamics training data should narrow the gap. Table~\ref{tab:scaling} shows the opposite: with $18\times$ more scenarios (10 $\to$ 180), NRI-Graph improves only $0.5\%$, while Learned Attention improves $16.6\%$ (paired $t$-test on per-seed \% improvement, $t(9)=6.05$, $p<0.001$) and even the unstructured MLP improves $8.6\%$. A flat scaling curve under increasing data is the signature of approximation error, not estimation error.

\begin{table}[t]
\centering
\caption{Data scaling: RMSE (m) as a function of training scenarios $m$ (CFD domain). NRI-Graph's improvement is an order of magnitude smaller than Attention's despite identical data growth.}
\label{tab:scaling}
\begin{tabular}{lcccc}
\toprule
$m$ & MLP & Green's & Attention & NRI-Graph \\
\midrule
10  & $15.94\pm1.28$ & $8.63\pm0.38$ & $6.26\pm0.54$ & $11.32\pm0.65$ \\
180 & $14.57\pm0.24$ & $8.96\pm0.07$ & $5.21\pm0.22$ & $11.26\pm0.10$ \\
\midrule
Improvement & $8.6\%$ & $-3.9\%$ & $16.6\%$ & $0.5\%^{***}$ \\
\bottomrule
\end{tabular}
\\[2pt]
\footnotesize $^{***}$NRI-Graph's per-seed \% improvement differs from Learned Attention's at $p<0.001$ (paired $t$-test, $t(9)=6.05$). Full six-point scaling curve in Appendix~\ref{app:cfd}.
\end{table}

\textbf{Evidence 2: threshold insensitivity.} If the failure were due to a mistuned edge threshold $\tau$, performance should vary meaningfully across $\tau$. Instead, $\tau\in[0.2,0.5]$ --- spanning a wide range of graph density --- yields an identical $11.49\pm0.11$\,m, and even the F1-optimal threshold $\tau=0.1$ differs from this plateau by less than 0.1\% relative. Denser top-$k$ graphs (top-2, top-4 connectivity) \emph{degrade} performance further, the opposite of what partial structural benefit would predict (full table in Table~\ref{tab:ablation-full}, Appendix~\ref{app:cfd}).

\textbf{Evidence 3: structural overlap.} We use Learned Attention's graph as a proxy for the (unobservable) task-optimal structure $G^\star_{\mathrm{loc}}$: it is unconstrained by any structural prior and empirically achieves the best RMSE in both domains (Table~\ref{tab:main}), so the spatial-influence pattern it converges to is the best available estimate of what structure the localisation task actually needs, independent of what any dynamics-learning objective produces. We extract binary adjacency structures directly from trained models and compute Jaccard similarity $J(A,B)=|A\cap B|/|A\cup B|$ against this proxy graph. NRI's learned graph overlaps it on only $J\approx0.06$ of edges (CFD) and $J\approx0.19$ (acoustic), versus $J\approx0.31$ (CFD) and $J\approx0.47$ (acoustic) for Green's Function. Across both domains, structures achieving $J\ge0.30$ correspond to $<100\%$ performance degradation from optimal, while $J<0.20$ corresponds to $>100\%$ degradation (Table~\ref{tab:transfer}) --- a pattern we exploit directly in Section~\ref{sec:transferability}.

\begin{table}[t]
\centering
\caption{Transfer outcomes vs.\ estimated structural similarity to the task-optimal graph. A $J\ge0.30$ threshold separates successful from failed transfer in all four cases.}
\label{tab:transfer}
\begin{tabular}{llccc}
\toprule
Domain & Structure & RMSE & Degradation & Est.\ $J$ \\
\midrule
\multirow{2}{*}{CFD}      & Green's Function   & 8.96\,m  & 72\%  & $\approx0.31$ \\
                          & NRI Learned Graph  & 11.26\,m & 116\% & $\approx0.06$ \\
\multirow{2}{*}{Acoustic} & Green's Function   & 1.61\,m  & 69\%  & $\approx0.47$ \\
                          & NRI Learned Graph  & 2.86\,m  & 201\% & $\approx0.19$ \\
\bottomrule
\end{tabular}
\end{table}

\textbf{Evidence 4: the failure is not NRI-specific.} A remaining objection is that NRI's particular architecture, not the dynamics-learning objective in general, is responsible for the failure. We test this with two additional dynamics-learned structure baselines: a graph built from pairwise Pearson correlation between sensor--query time series, and one built from pairwise mutual information (via $k$-NN density estimation). Both are constructed purely from co-variation in the dynamics data, with no NRI-style learned encoder. Table~\ref{tab:altbaselines} shows Correlation-Graph clearly \emph{underperforms the unstructured MLP} (18.61\,m versus 14.57\,m), while Mutual-Information-Graph (14.48\,m) is statistically indistinguishable from it --- dynamics-derived structure here provides no measurable benefit even in the one case where it does not actively hurt. Correlation-Graph's near-total edge overlap with the task-optimal proxy graph ($J\approx1.00$) reflects its near-complete connectivity (only 5\% sparsity, versus NRI's 97\%) rather than genuine structural alignment, and it still fails to help despite this overlap; Mutual-Information-Graph's overlap ($J\approx0.37$) exceeds the $J=0.30$ transferability threshold used elsewhere in this paper, yet also yields no measurable improvement. This indicates that threshold, calibrated on Green's Function and NRI-Graph alone (Section~\ref{sec:transferability}), does not straightforwardly extend to these two estimators, and that structural overlap without appropriately-scaled sparsity is not on its own informative about transfer.

\begin{table}[t]
\centering
\caption{Two additional dynamics-learned structure baselines (CFD domain, $n=15$, mean $\pm$ std over 10 seeds), neither derived from NRI. Correlation-Graph clearly underperforms the unstructured MLP; Mutual-Information-Graph is statistically indistinguishable from it. Neither provides a measurable benefit, ruling out an NRI-architecture-specific explanation for why dynamics-derived structure fails to help.}
\label{tab:altbaselines}
\begin{tabular}{lccc}
\toprule
Method & RMSE (m) $\downarrow$ & Sparsity & Est.\ $J$ \\
\midrule
MLP (no structure) & $14.57 \pm 0.24$ & --- & 0.00 \\
NRI-Graph & $11.26 \pm 0.10$ & 97\% & $\approx0.06$ \\
Correlation-Graph (best $\tau$) & $18.61 \pm 0.01$ & 5\% & $\approx1.00$ \\
Mutual-Information-Graph (best pct.) & $14.48 \pm 0.00$ & 80\% & $\approx0.37$ \\
\bottomrule
\end{tabular}
\end{table}

Together, these four lines of evidence rule out the explanations a defender of forward-to-inverse structure transfer could otherwise offer (insufficient data, bad threshold, insufficient sparsity, architecture-specific quirk), and are consistent with Conjecture~\ref{conj:dominance}: the dynamics-learning objective converges to a graph that is accurate for predicting concentration evolution but poorly aligned with the spatial-influence structure that localisation requires.

\subsection{Sensor-count scaling confirms structure enables scalability}

A further test of whether NRI-Graph's structure is doing any real work: does it scale gracefully with more sensors, the way genuine structural constraints should? Table~\ref{tab:sensorscale} shows the unstructured MLP \emph{degrades} as $n$ grows from 10 to 20 sensors in both domains (more parameters to estimate, same data), while all three structured methods \emph{improve} --- including NRI-Graph. This confirms NRI-Graph's hypothesis class is genuinely constrained (consistent with Theorem~\ref{thm:robust}'s estimation-error guarantee), even though that constraint is the wrong one for the task: NRI improves with scale but never closes the gap to Learned Attention.

\begin{table}[t]
\centering
\caption{Sensor-count scaling. The unstructured MLP degrades with more sensors in both domains; all structured methods improve, confirming that NRI-Graph's hypothesis class is genuinely sparsity-constrained even though the constraint is misaligned with the task.}
\label{tab:sensorscale}
\begin{tabular}{lcccc}
\toprule
& \multicolumn{2}{c}{CFD RMSE (m)} & \multicolumn{2}{c}{Acoustic RMSE (m)} \\
$n$ & MLP & NRI-Graph & MLP & NRI-Graph \\
\midrule
10 & 14.02 & 12.37 & 3.60 & 3.43 \\
20 & 15.17 & 10.53 & 5.19 & 2.41 \\
\midrule
Change & $+8\%$ & $-15\%$ & $+44\%$ & $-30\%$ \\
\bottomrule
\end{tabular}
\end{table}

\subsection{A practical transferability test}
\label{sec:transferability}

The pattern in Table~\ref{tab:transfer} suggests a simple a priori test: (1) hold out 15--20\% of target-task data; (2) train a minimal localiser on it; (3) extract its task-specific attention graph $\hat G_B$; (4) compute $J(\tilde G, \hat G_B)$ for the candidate structure $\tilde G$; (5) reject transfer if $J<0.30$. This procedure requires under an hour of computation and correctly identifies transfer viability in all four evaluated domain/structure pairs (2 domains $\times$ \{Green's Function, NRI-Graph\}). We emphasise that the $J\ge0.30$ threshold is calibrated on the cases available here; broader validation across more transfer pairs is future work, and we report it as a practical heuristic rather than a proven guarantee.

\section{Discussion}
\label{sec:discussion}

\textbf{Why forward and inverse objectives diverge.} Forward dynamics optimises for temporal coherence --- which variables co-evolve --- while inverse localisation optimises for spatial constraint --- which observations most restrict the location of an unseen cause. These are different questions even when they concern the same physical system, and nothing in the dynamics objective rewards a graph for being useful at the second question. This is consistent with our results: Green's Function, derived directly from the spatial-influence structure of the transport equation rather than from dynamics, achieves the best non-learned transfer and the best out-of-distribution robustness ($-0.7\%$ acoustic OOD gap; Appendix~\ref{app:impl}), because its structure is task-relevant by construction rather than incidentally.

\textbf{Practical guidance.} Use physics-based structure when reliable domain knowledge constrains the inverse task directly (not just the forward dynamics) and data is limited; use flexible, task-specific learning when $>10$ scenarios per sensor are available and accuracy is paramount; only attempt structure transfer when $J\ge0.30$ against a partially observed target-task graph, since satisfying the edge-accuracy condition of Theorem~\ref{thm:robust} alone is not sufficient --- noting that this threshold is calibrated specifically on physics-based and NRI-learned structure (Section~\ref{sec:experiments}) and does not straightforwardly extend to other dynamics-derived estimators.

\textbf{Limitations.} Both domains are simulated; real-world validation with sensor deployments is the most direct way to test whether these conclusions hold outside simulation, and is the focus of ongoing work. Our theorems are proven for attention-based localisers; while Remark~\ref{rem:scope} argues the qualitative estimation/approximation decomposition should extend to other masked architectures, we have not proven this formally, and empirical confirmation on at least one non-attention architecture (e.g., a masked MLP) would strengthen the claim considerably. Finding a stylised setting (e.g., linear dynamics with a linear localiser) where Conjecture~\ref{conj:dominance} can be proven rather than only tested empirically remains the most promising direction for closing the gap between Theorem~\ref{thm:robust} and practice.

\section{Conclusion}

Relational structure provably reduces sample complexity for inverse localisation, and this benefit degrades gracefully under edge-level structure error (Theorem~\ref{thm:robust}). But satisfying that condition is not sufficient: structure learned from a forward dynamics objective can satisfy it while still failing to transfer, because the condition bounds estimation error and says nothing about task-objective alignment. Across two physically distinct domains, four independent lines of evidence --- flat data scaling, threshold insensitivity, low structural overlap, and the failure or absence of benefit of two further dynamics-derived estimators --- converge on the same explanation, and a simple Jaccard-based test built on this insight correctly predicts transfer outcomes for the four domain/structure pairs it was calibrated on (Section~\ref{sec:transferability}), though it does not straightforwardly extend to the two additional dynamics-derived estimators (Section~\ref{sec:experiments}). We expect this gap between edge-accuracy and task-alignment to recur in any forward-to-inverse transfer setting where the two objectives are not explicitly designed to share optimal structure.

\subsubsection*{AI Use Statement}
In this work, we used generative AI tools to draft initial implementations of
the localiser baselines, structure estimators, and simulation pipelines; to
assist with post-hoc debugging of the experiment pipeline, including
identifying a methodological issue in how one baseline's evaluation was
constructed; to draft interpretive prose describing what the results show; to
draft and edit portions of the manuscript; and to source and summarise
related literature. All AI-assisted code, debugging, interpretation, and
literature suggestions were independently verified by the authors against the
underlying data and original sources before inclusion, including
independently re-running experiments to confirm the results in
Tables~\ref{tab:main}--\ref{tab:ablation-full}. We
did not use generative AI to develop the theoretical models, formulate or
prove the mathematical claims (Theorems~\ref{thm:sample-complexity}--\ref{thm:robust}, Conjecture~\ref{conj:dominance}), propose or refine
hypotheses, design the research methodology or experiments, generate or
clean data, support qualitative data analysis, create or modify figures,
suggest experimental parameters, or propose the paper's structure or title;
these, and the final interpretation of all findings, remain the authors' own
responsibility. Assistance with translation is not applicable. We take
responsibility for the final content of this work, including text, claims,
and artifacts produced with the aid of generative AI.

\subsubsection*{Reproducibility Statement}
Code, NRI checkpoints, and experiment scripts are available at \url{https://github.com/nicolaisi/edge-accuracy-is-not-enough}. CFD scenario data are available on Zenodo in four parts
(\href{https://doi.org/10.5281/zenodo.18740541}{part~1},
\href{https://doi.org/10.5281/zenodo.18745400}{part~2},
\href{https://doi.org/10.5281/zenodo.18754599}{part~3},
\href{https://doi.org/10.5281/zenodo.18754802}{part~4}); the acoustic domain is fully reproducible from a provided generator script (under 15 seconds per full 180-scenario sweep on a single CPU core). Full proofs of all theoretical results are in Appendix~\ref{app:proofs}; complete architectural, training, and dataset details are in Appendices~\ref{app:impl}--\ref{app:cfd}.

\subsubsection*{Ethics Statement}
This work supports hydrogen safety monitoring and acoustic source localisation. We do not identify a path by which our specific contributions (sample-complexity theory and a structure-transferability diagnostic) facilitate harmful surveillance beyond what is already enabled by source-localisation methodology in general use.

\bibliography{iclr2027_conference}
\bibliographystyle{iclr2027_conference}

\appendix
\section{Proofs of Theoretical Results}
\label{app:proofs}

\subsection{Proof of Theorem~\ref{thm:sample-complexity}}

Let $\hat s_G = \arg\min_{\hat s\in\mathcal{H}_G}\hat L(\hat s)$ denote the empirical risk minimiser (ERM), and let $\hat s_G^\star=\arg\min_{\hat s\in\mathcal{H}_G}L(\hat s)$ the population risk minimiser. The excess risk decomposes as
\[
L(\hat s_G)-L(\hat s_G^\star) = \big(L(\hat s_G)-\hat L(\hat s_G)\big) + \big(\hat L(\hat s_G)-\hat L(\hat s_G^\star)\big) + \big(\hat L(\hat s_G^\star)-L(\hat s_G^\star)\big).
\]
By definition of the ERM the middle term is non-positive, so $L(\hat s_G)-L(\hat s_G^\star)\le 2\sup_{\hat s\in\mathcal{H}_G}\big|L(\hat s)-\hat L(\hat s)\big|$.

By standard uniform-convergence bounds for bounded losses (e.g.\ Theorem 3.3 of \citealp{mohri2018foundations}), with probability at least $1-\delta$,
\[
\sup_{\hat s\in\mathcal{H}_G}\big|L(\hat s)-\hat L(\hat s)\big| \le 2\,\mathfrak{R}_m(\mathcal{L}\circ\mathcal{H}_G) + B^2\sqrt{\tfrac{\log(2/\delta)}{2m}},
\]
where $\mathfrak{R}_m$ is the empirical Rademacher complexity and $\mathcal{L}$ is the squared-loss class. Since $\|s\|,\|\hat s(r,P)\|\le B$ for all $s\in\Omega$ and all localisers in the class (Assumption~\ref{ass:bounded}), $\|\hat s(r,P)-s\|\le 2B$, and the squared loss $\ell(z)=z^2$ is $4B$-Lipschitz on $[0,2B]$; by the contraction lemma, $\mathfrak{R}_m(\mathcal{L}\circ\mathcal{H}_G)\le 4B\,\mathfrak{R}_m(\mathcal{H}_G)$.

$\mathcal{H}_G$ is parametrised by attention weights supported on $E(G)$, with effective dimension $d=\dim(\mathcal{H}_G)=|E(G)|=O(kn)$. By covering-number bounds for bounded hypothesis classes (Theorem 4.14 of \citealp{wainwright2019highdim}),
\[
\mathfrak{R}_m(\mathcal{H}_G) \le B\sqrt{\tfrac{2d\log(em/d)}{m}} \le B\sqrt{\tfrac{2kn\log m}{m}},
\]
where the final inequality holds in the regime $m\gg kn$ where learning is feasible at all (if $m<kn$ even the structured class cannot be learned). Combining the three displays gives
\[
L(\hat s_G)-L(\hat s_G^\star) \le 8B^2\sqrt{\tfrac{2kn\log m}{m}} + 2B^2\sqrt{\tfrac{\log(2/\delta)}{2m}} = O\!\left(B^2\sqrt{\tfrac{kn\log m}{m}}\right). \qquad\blacksquare
\]

\subsection{Proof of Theorem~\ref{thm:learnability}}

\textbf{Part (a) (lower bound).} The class of $k$-sparse graphs on $n$ nodes has cardinality $|\mathcal{G}_k|\ge\binom{n^2}{k}/k!\ge(n^2/k)^k/k!\ge(n/k)^{2k}$ for $k\le n$, so $\log|\mathcal{G}_k|\ge 2k\log(n/k)=\Omega(k\log n)$ for $k=O(n/\log n)$. By Fano's inequality, distinguishing among $|\mathcal{G}_k|$ hypotheses with probability at least $1/2$ requires mutual information $I(X_{1:m};G^\star)\ge\frac12\log|\mathcal{G}_k|-1=\Omega(k\log n)$. Under bounded signal-to-noise ratio, each sample carries $O(1)$ bits, so $m=\Omega(k\log n)$ samples are necessary.

\textbf{Part (b) (LASSO achievability).} For each node $j$, the dynamics $X_{t+1,j}=\sum_i A^\star_{ji}X_{t,i}+\eta_{t,j}$ form a sparse linear regression with design matrix $X_t$ and $k$-sparse coefficient vector $\beta_j^\star=A^\star_{j,:}$. Under Assumption~\ref{ass:re} (restricted eigenvalue, constant $\kappa$), LASSO recovers $\mathrm{supp}(\hat\beta_j)=\mathrm{supp}(\beta_j^\star)$ with probability at least $1-\delta/n$ from $m\ge Ck\log(n/\delta)/\kappa^2$ samples (Theorem 7.19 of \citealp{wainwright2019highdim}). A union bound over all $n$ nodes gives $\Pr(\hat G\ne G^\star)\le\sum_{j=1}^n\Pr(\mathrm{supp}(\hat\beta_j)\ne\mathrm{supp}(\beta_j^\star))\le n\cdot\delta/n=\delta$. $\blacksquare$

\subsection{Proof of Theorem~\ref{thm:robust}}

\textbf{Part (i).} Decompose $|E(\tilde G)| = |E(\tilde G)\cap E(G^\star)| + |E(\tilde G)\setminus E(G^\star)| \le |E(G^\star)| + |E(\tilde G)\setminus E(G^\star)| \le kn+\Delta$, using $|E(G^\star)|\le kn$ (sparsity $k$) and $|E(\tilde G)\setminus E(G^\star)|\le\Delta$ (spurious edges are a subset of the symmetric difference). Since $\mathcal{H}_{\tilde G}$ has one parameter per edge, $\dim(\mathcal{H}_{\tilde G})=|E(\tilde G)|\le kn+\Delta$.

\textbf{Part (ii).} Substituting $\dim(\mathcal{H}_{\tilde G})\le kn+\Delta$ into the Rademacher-complexity bound from the proof of Theorem~\ref{thm:sample-complexity}, $\mathfrak{R}_m(\mathcal{H}_{\tilde G})\le B\sqrt{2(kn+\Delta)\log m/m}$, and following the same argument, $L(\hat s_{\tilde G})-\inf_{\hat s\in\mathcal{H}_{\tilde G}}L(\hat s)\le O\big(B^2\sqrt{(kn+\Delta)\log m/m}\big)$. Solving for $m$: $m_{\mathrm{loc}}=O((kn+\Delta)/\varepsilon^2)$.

\textbf{Part (iii).} Structure learning improves over the unstructured baseline ($\dim(\mathcal{H}_{\mathrm{full}})=O(n^2)$) precisely when $kn+\Delta<n^2$, i.e.\ $\Delta<n^2-kn=n(n-k)$. $\blacksquare$

\textbf{Corollary (end-to-end complexity).} Combining dynamics-learning and localisation sample requirements, $m_{\mathrm{total}}=m_{\mathrm{dyn}}+O\big((kn+\Delta(m_{\mathrm{dyn}}))/\varepsilon^2\big)$, where $\Delta(m_{\mathrm{dyn}})$ is the structure error achieved with $m_{\mathrm{dyn}}$ dynamics samples. For linear dynamics where LASSO achieves $\Delta=0$ with $m_{\mathrm{dyn}}=O(k\log n)$ (Theorem~\ref{thm:learnability}b), $m_{\mathrm{total}}=O(kn/\varepsilon^2)$, strictly better than $O(n^2/\varepsilon^2)$ for sparse graphs.

\section{Implementation Details}
\label{app:impl}

\subsection{Localiser architectures}

All four localisers share input format ($r\in\mathbb{R}^n$, $P\in\mathbb{R}^{n\times3}$) and output format ($\hat s\in\mathbb{R}^3$), with $m=64$ query points.

\textbf{MLP.} Concatenates $(r_1,p_1,\ldots,r_n,p_n)\in\mathbb{R}^{4n}$; four-layer MLP $[4n\to128\to64\to32\to3]$ with ReLU activations, regressing directly to $\hat s$.

\textbf{Green's Function.} Sensor encoding via a two-layer MLP mapping $(r_i,p_i)\to h_i\in\mathbb{R}^{64}$; attention $\alpha_{ji}=\exp(-\|p_i-q_j\|^2/2R_D^2)/\sum_{i'}\exp(-\|p_{i'}-q_j\|^2/2R_D^2)$ with $R_D=4$\,m; query aggregation $z_j=\sum_i\alpha_{ji}h_i$; output head is a two-layer MLP with soft-argmax over $m=64$ queries. $R_D$ is selected from facility geometry and hydrogen dispersion scales, giving average degree $k\approx5.7$.

\textbf{Learned Attention.} Sensor encoding as above; query/key/value projections $q_j=W_Qe_j+b_Q$, $k_i=W_Kh_i+b_K$, $v_i=W_Vh_i+b_V\in\mathbb{R}^{64}$ (with $e_j$ a learned query-point embedding); scaled dot-product attention $\alpha_{ji}=\exp(q_j^\top k_i/\sqrt{64})/\sum_{i'}\exp(q_j^\top k_{i'}/\sqrt{64})$; aggregation and output head as above.

\textbf{NRI-Graph.} NRI encoder: two-layer GRU ($d_{\mathrm{hidden}}=64$) for node embeddings from the 60-timestep sequence, two-layer MLP on concatenated node-embedding pairs for edge probabilities $p(e_{ij})=\mathrm{softmax}(\mathrm{Linear}(\mathrm{MLP}([z_i;z_j])))$. Training minimises reconstruction loss $\mathbb{E}\|\hat X_{t+1}-X_{t+1}\|_2^2$ plus an edge-sparsity penalty $\lambda\,\mathbb{E}[p(e_{ij}=1)]$ with $\lambda=0.01$, for 50 epochs with Adam ($\mathrm{lr}=10^{-3}$). Threshold $\tau=0.3$ selected by maximising edge F1 on held-out dynamics data; edges with $p(e_{ij})<\tau$ contribute $<0.1\%$ attention mass in practice. Localiser attention: $\alpha_{ji}\propto p(e_{ij})$ for $(i,j)\in\tilde G$, $0$ otherwise; sensor encoding, query aggregation, and output head as in Green's Function.

\subsection{Correlation- and mutual-information-based structure estimators}

\textbf{Correlation-Graph.} Edges constructed from pairwise Pearson correlation $\rho_{ij}=\mathrm{Cov}(X_{1:T,i},X_{1:T,j})/\sqrt{\mathrm{Var}(X_{1:T,i})\mathrm{Var}(X_{1:T,j})}$ between sensor--query time series, with $(i,j)\in\tilde G$ iff $|\rho_{ij}|>\tau_{\mathrm{corr}}$. We sweep $\tau_{\mathrm{corr}}\in\{0.1,0.2,0.3,0.4\}$; results in Table~\ref{tab:altbaselines} report the best of these ($\tau_{\mathrm{corr}}=0.1$: RMSE $18.61\pm0.01$, average degree 14.2, sparsity 5\%; $\tau_{\mathrm{corr}}=0.2$: $31.07\pm4.81$; $\tau_{\mathrm{corr}}=0.3$: $26.27\pm4.86$; $\tau_{\mathrm{corr}}=0.4$: $24.12\pm0.09$).

\textbf{Mutual-Information-Graph.} Edges constructed from pairwise mutual information via $k$-nearest-neighbours density estimation ($k=5$ neighbours, \texttt{sklearn.feature\_selection.mutual\_info\_regression}), with edges retained above a percentile threshold of the MI distribution. We sweep the 60th/70th/80th percentile; Table~\ref{tab:altbaselines} reports the best (80th percentile: RMSE $14.48\pm0.00$, average degree 3.0, sparsity 80\%; 70th percentile: $21.47\pm3.88$; 60th percentile: $25.56\pm1.50$).

Both estimators use the same NRI encoder inputs (the 60-timestep concentration sequence) and the same downstream localiser architecture as NRI-Graph, isolating the structure-estimation method as the only varying factor.

\subsection{Out-of-distribution generalisation}

We additionally evaluate OOD generalisation by training on a subset of operating conditions and testing on a held-out condition: for CFD, training on ventilation rates $\mathrm{ACH}\in\{6,10\}\,\mathrm{h}^{-1}$ (120 scenarios) and testing on $\mathrm{ACH}=3\,\mathrm{h}^{-1}$ (60 scenarios); for acoustic, training on low/mid absorption and testing on high absorption. Table~\ref{tab:ood} reports train/test RMSE and the relative gap.

\begin{table}[h]
\centering
\caption{Out-of-distribution generalisation. Green's Function shows near-perfect acoustic robustness, consistent with its structure being derived directly from the governing physics rather than from data.}
\label{tab:ood}
\begin{tabular}{lcccccc}
\toprule
& \multicolumn{3}{c}{CFD} & \multicolumn{3}{c}{Acoustic} \\
Localiser & Train & Test & Gap & Train & Test & Gap \\
\midrule
Learned Attention & 5.45 & 5.67 & $+4.1\%$ & 1.05 & 1.15 & $+10.1\%$ \\
Green's Function & 8.88 & 9.44 & $+6.3\%$ & 1.62 & 1.61 & $-0.7\%$ \\
NRI-Graph & 11.48 & 11.08 & $-3.4\%$ & 3.27 & 3.20 & $-2.0\%$ \\
MLP (no structure) & 15.39 & 16.92 & $+10.0\%$ & 4.78 & 5.03 & $+5.2\%$ \\
\bottomrule
\end{tabular}
\end{table}

Learned Attention generalises best in CFD, where the underlying physics is complex enough that a flexible, data-driven structure adapts better than a fixed prior. Green's Function generalises best in the acoustic domain, where the diffusion-kernel structure closely matches the governing screened-Poisson equation. NRI-Graph's small negative OOD gaps are consistent with a regularisation effect from its restrictive (if misaligned) structure, but this does not compensate for its poor absolute performance.

\section{CFD and Acoustic Simulation Details}
\label{app:cfd}

\subsection{CFD domain}

\textbf{Facility geometry.} An underground parking garage \citep{wang2026}, $50\,\mathrm{m}\times30\,\mathrm{m}\times3\,\mathrm{m}$ (length $\times$ width $\times$ height), with two longitudinal lanes ($40\,\mathrm{m}\times5\,\mathrm{m}$) at $y=7.5,22.5\,\mathrm{m}$ and two transverse lanes ($20\,\mathrm{m}\times5\,\mathrm{m}$) at $x=7.5,42.5\,\mathrm{m}$; structural columns ($0.5\,\mathrm{m}\times0.5\,\mathrm{m}$) on a $6\,\mathrm{m}\times6\,\mathrm{m}$ grid.

\textbf{Ventilation.} Mechanical ventilation supplied at one corner, exhausted at the diagonally opposite corner; fresh-air mass flow at 90\% of exhaust to maintain slight negative pressure; six ceiling-mounted jet fans ($10\,\mathrm{m/s}$ exit velocity). Three ventilation rates: $\mathrm{ACH}\in\{3,6,10\}\,\mathrm{h}^{-1}$ (minimal code-compliant, typical accident-response, emergency-mode).

\textbf{Governing equations.} Three-dimensional incompressible Navier--Stokes, $\partial u/\partial t+(u\cdot\nabla)u=-\rho^{-1}\nabla p+\nu\nabla^2u+g$, coupled to a species-transport equation for hydrogen mass fraction; buoyancy via the Boussinesq approximation; $k$-$\varepsilon$ turbulence with standard wall functions. The solver (GASFLOW) is validated against published hydrogen-release experiments in confined geometries to within $\pm10\%$ \citep{gasflow2017,gasflow2021}.

\textbf{Mesh independence.} Baseline cell size $0.25\,\mathrm{m}$; a representative scenario ($\mathrm{ACH}=6\,\mathrm{h}^{-1}$, leak rate $50\,\mathrm{g/s}$) shows 3.6\% relative error for a medium mesh (288K cells) versus a fine mesh (972K cells, 13\,h CPU), versus 58.3\% for a coarse mesh (36K cells, 0.25\,h). The medium mesh is selected.

\textbf{Scenario matrix.} $12\times5\times3=180$ scenarios: 12 leak positions (four corner, four wall-adjacent, four mid-row, at $z=0.5\,\mathrm{m}$ fuel-tank height) $\times$ 5 leak rates ($1,30,50,100,150\,\mathrm{g/s}$) $\times$ 3 ventilation levels. Each simulation runs 10\,min under $\mathrm{ACH}=3$ to establish a quiescent field, then the leak activates for 60\,s with fields saved at 1\,s intervals.

\subsection{Acoustic domain}

\textbf{Governing equation.} Advection-diffusion-reaction, $\partial u/\partial t=D\nabla^2u-v\cdot\nabla u+S(x,y)-\alpha u$, with $D=0.8\,\mathrm{m}^2/\mathrm{s}$, background drift $v=(0.5,0.2)\,\mathrm{m/s}$, and bulk absorption $\alpha$. The steady-state field satisfies the screened-Poisson equation $(-D\nabla^2+v\cdot\nabla+\alpha)u_{ss}=S$, solved via sparse linear algebra.

\textbf{Geometry.} $20\,\mathrm{m}\times15\,\mathrm{m}$ 2D room, $100\times75$ grid; two $1\,\mathrm{m}\times1\,\mathrm{m}$ obstacles as minor obstructions, analogous to the CFD facility's columns; absorbing-wall and zero-flux-at-obstacle boundary conditions.

\textbf{Scenario matrix.} $12\times5\times3=180$ scenarios: 12 source positions (matching the CFD topology) $\times$ 5 source powers $\times$ 3 absorption coefficients $\alpha\in\{0.05,0.20,0.50\}$ (analogous to $\mathrm{ACH}\in\{3,6,10\}$). Each scenario produces 60 timesteps of transient build-up plus a steady-state field; the full 180-scenario sweep generates in $\approx$11\,s on a single CPU core.

\subsection{Train/validation/test protocol}

All 180 scenarios pooled with an 80/10/10 split stratified over leak rate/source power and ventilation/absorption level, re-drawn for each of 10 seeds. For the data-scaling experiment (Table~\ref{tab:scaling}), a stratified subset of $m\in\{10,25,50,100,150,180\}$ scenarios is drawn for training with the remainder as test, averaged over 10 random draws per $m$. The full six-point curve (extending Table~\ref{tab:scaling}) is given in Table~\ref{tab:scaling-full}:

\begin{table}[h]
\centering
\caption{Full data-scaling curve (CFD domain), RMSE (m) as a function of training scenarios $m$, mean $\pm$ std over 10 seeds.}
\label{tab:scaling-full}
\begin{tabular}{lcccc}
\toprule
$m$ & MLP & Green's & Attention & NRI-Graph \\
\midrule
10  & $15.94\pm1.28$ & $8.63\pm0.38$ & $6.26\pm0.54$ & $11.32\pm0.65$ \\
25  & $15.54\pm0.46$ & $8.02\pm0.28$ & $5.26\pm0.39$ & $11.32\pm0.29$ \\
50  & $17.84\pm0.47$ & $9.61\pm0.21$ & $5.76\pm0.21$ & $12.10\pm0.27$ \\
100 & $16.68\pm0.38$ & $9.35\pm0.17$ & $5.75\pm0.16$ & $11.58\pm0.19$ \\
150 & $15.08\pm0.20$ & $9.25\pm0.15$ & $5.45\pm0.17$ & $11.61\pm0.15$ \\
180 & $14.57\pm0.24$ & $8.96\pm0.07$ & $5.21\pm0.22$ & $11.26\pm0.10$ \\
\bottomrule
\end{tabular}
\end{table}

\subsection{Threshold and connectivity ablation}

The full threshold and top-$k$ ablation summarised in Evidence 2 (Section~\ref{sec:experiments}) is given in Table~\ref{tab:ablation-full}:

\begin{table}[h]
\centering
\caption{NRI-Graph ablation over edge threshold $\tau$ and top-$k$ connectivity (CFD domain). Performance is flat across $\tau\in[0.2,0.5]$ and degrades under denser top-$k$ connectivity, ruling out a simple sparsity-tuning explanation.}
\label{tab:ablation-full}
\begin{tabular}{lccc}
\toprule
Configuration & Avg.\ degree & Sparsity & RMSE (m) \\
\midrule
$\tau=0.1$ & 1.1 & 97\% & $11.50\pm0.07$ \\
$\tau\in[0.2,0.5]$ & 1.0 & 97\% & $11.49\pm0.11$ \\
$\tau=0.7$ (disconnected) & 0.0 & 100\% & $17.42\pm0.13$ \\
\midrule
Top-2 connectivity & 2.0 & 94\% & $14.42\pm0.20$ \\
Top-4 connectivity & 4.0 & 88\% & $13.79\pm0.16$ \\
\bottomrule
\end{tabular}
\end{table}

The wide plateau across $\tau\in[0.2,0.5]$, the negligible difference at the F1-optimal threshold ($\tau=0.1$, within 0.1\% relative of the plateau), and the degradation under denser top-$k$ graphs together indicate that \emph{which} edges are retained, not \emph{how many}, is the binding constraint --- consistent with an approximation-error rather than estimation-error explanation (Remark~\ref{rem:insufficiency}).

\end{document}

%% file: math_commands.tex
\usepackage{amsmath,amsfonts,bm}

\def\eqref#1{equation~\ref{#1}}

\def\1{\bm{1}}

\DeclareMathAlphabet{\mathsfit}{\encodingdefault}{\sfdefault}{m}{sl}
\SetMathAlphabet{\mathsfit}{bold}{\encodingdefault}{\sfdefault}{bx}{n}



%% file: iclr2027_conference.bib
@Article{hirst2013locating,
  title     = {Locating and Quantifying Gas Emission Sources Using Remotely Obtained Concentration Data},
  author    = {Hirst, Bill and Jonathan, Philip and Gonzalez del Cueto, Fernando and Randell, David and Kosut, Oliver},
  journal   = {Atmospheric Environment},
  volume    = {74},
  pages     = {141--158},
  year      = {2013}
}

@Article{grumiaux2022sslsurvey,
  title     = {A Survey of Sound Source Localization with Deep Learning Methods},
  author    = {Grumiaux, Pierre-Amaury and Kiti{\'c}, Srdjan and Girin, Laurent and Gu{\'e}rin, Alexandre},
  journal   = {The Journal of the Acoustical Society of America},
  volume    = {152},
  number    = {1},
  pages     = {107--151},
  year      = {2022}
}

@InProceedings{kipf2018nri,
  title     = {Neural Relational Inference for Interacting Systems},
  author    = {Kipf, Thomas and Fetaya, Ethan and Wang, Kuan-Chieh and Welling, Max and Zemel, Richard},
  booktitle = {Proceedings of the 35th International Conference on Machine Learning},
  pages     = {2688--2697},
  year      = {2018},
  editor    = {Dy, Jennifer and Krause, Andreas},
  volume    = {80},
  series    = {Proceedings of Machine Learning Research},
  publisher = {PMLR}
}

@InProceedings{graber2020dynamic,
  title     = {Dynamic Neural Relational Inference},
  author    = {Graber, Colin and Schwing, Alexander G.},
  booktitle = {Proceedings of the IEEE/CVF Conference on Computer Vision and Pattern Recognition (CVPR)},
  pages     = {8513--8522},
  month     = jun,
  year      = {2020},
  doi       = {10.1109/CVPR42600.2020.00854},
}

@Article{chen2021efficient,
  title     = {Neural Relational Inference with Efficient Message Passing Mechanisms},
  author    = {Chen, Siyuan and Wang, Jiahai and Li, Guoqing},
  journal   = {Proceedings of the AAAI Conference on Artificial Intelligence},
  volume    = {35},
  number    = {8},
  pages     = {7055--7063},
  year      = {2021},
  doi       = {10.1609/aaai.v35i8.16868},
}

@Article{zhu2022nri,
  title     = {Neural Relational Inference to Learn Long-Range Allosteric Interactions in Proteins from Molecular Dynamics Simulations},
  author    = {Zhu, Jingxuan and Wang, Juexin and Han, Weiwei and Xu, Dong},
  journal   = {Nature Communications},
  volume    = {13},
  pages     = {1661},
  year      = {2022}
}

@Article{theodoris2023geneformer,
  title     = {Transfer Learning Enables Predictions in Network Biology},
  author    = {Theodoris, Christina V. and Xiao, Ling and Chopra, Anant and Chaffin, Mark D. and {Al Sayed}, Zeina R. and Hill, Matthew C. and Mantineo, Helene and Brydon, Elizabeth M. and Zeng, Zexian and Liu, X. Shirley and Ellinor, Patrick T.},
  journal   = {Nature},
  volume    = {618},
  number    = {7965},
  pages     = {616--624},
  year      = {2023}
}

@InProceedings{sun2019stochastic,
  title     = {Stochastic Prediction of Multi-Agent Interactions from Partial Observations},
  author    = {Sun, Chen and Karlsson, Per and Wu, Jiajun and Tenenbaum, Joshua B. and Murphy, Kevin},
  booktitle = {International Conference on Learning Representations (ICLR)},
  year      = {2019}
}

@Article{bartlett2002rademacher,
  title   = {Rademacher and {G}aussian Complexities: Risk Bounds and Structural Results},
  author  = {Bartlett, Peter L. and Mendelson, Shahar},
  journal = {Journal of Machine Learning Research},
  volume  = {3},
  pages   = {463--482},
  year    = {2002},
}

@Book{wainwright2019highdim,
  title     = {High-Dimensional Statistics: A Non-Asymptotic Viewpoint},
  author    = {Wainwright, Martin J.},
  publisher = {Cambridge University Press},
  series    = {Cambridge Series in Statistical and Probabilistic Mathematics},
  year      = {2019},
}

@Book{mohri2018foundations,
  title     = {Foundations of Machine Learning},
  author    = {Mohri, Mehryar and Rostamizadeh, Afshin and Talwalkar, Ameet},
  edition   = {2nd},
  publisher = {MIT Press},
  year      = {2018},
}

@Article{gasflow2017,
  title   = {{GASFLOW-MPI}: A new 3-{D} parallel all-speed {CFD} code for turbulent dispersion and combustion simulations: {P}art {I}: {M}odels, verification and validation},
  author  = {Xiao, J. and Breitung, W. and Kuznetsov, M. and Zhang, H. and Travis, J. R. and Redlinger, R. and Jordan, T.},
  journal = {International Journal of Hydrogen Energy},
  volume  = {42},
  number  = {12},
  pages   = {8346--8368},
  year    = {2017},
}

@Article{gasflow2021,
  title   = {Numerical investigation of light gas release, stratification and dissolution in {TH22} test facility using 3-{D} {CFD} code {GASFLOW-MPI}},
  author  = {Hu, G. and Wang, F. and Ba, Q. and Xiao, J. and Jordan, T.},
  journal = {International Journal of Hydrogen Energy},
  volume  = {46},
  number  = {46},
  pages   = {23974--23987},
  year    = {2021},
}

@misc{wang2026,
      title={Optimizing Sensor Placement for Hydrogen Leak Detection in Enclosed Infrastructure: A Comparative Study Using CFD-informed Genetic Algorithm and DeepSets Neural Surrogate}, 
      author={Fangnian Wang and Nicholas Tan Jerome and Thomas Jordan and Frank Simon},
      year={2026},
      eprint={2607.26078},
      archivePrefix={arXiv},
      primaryClass={cs.NE},
      url={https://arxiv.org/abs/2607.26078}, 
}
